\documentclass{itor}

\usepackage{natbib}%
\usepackage[figuresright]{rotating}

\usepackage{url}
\usepackage{amsfonts}
\usepackage[ruled, linesnumbered]{algorithm2e}
\usepackage[normalem]{ulem}
\usepackage{xcolor}
\usepackage{color}
\usepackage{subcaption}
\usepackage{multirow}
\usepackage{listings}
\usepackage{comment}
\usepackage{enumerate}
\usepackage{amsmath,amsthm}
\usepackage{braket}

\usepackage{tikz}
\usetikzlibrary{trees}
\theoremstyle{definition}

\theoremstyle{remark}
\newtheorem{definition}{Definition}

\newcommand{\yc}{\textcolor{purple}}
\newcommand{\nat}{\textcolor{blue}}

\newcommand{\nouveau}{\textcolor{black}}

\usepackage{environ}

\NewEnviron{multisout}{%
  \protected@edef\tmp{\BODY}%
  \expandafter\domultisout\BODY\par\relax
}
\makeatletter
\def\domultisout#1\par#2{%
  \sout{#1}%
  \ifx\relax#2\else
    \par
    \expandafter\domultisout\expandafter#2%
  \fi
}
\makeatother

\jname{International Transactions in Operational Research}
\jvol{XX}
\jyear{20XX}
\doi{xx.xxxx/itor.xxxxx}

\begin{document}

\title{A decision-support system applied to Law: Reasoning and explainability of the decision}

\author[Jeremy BOUCHE-PILLON]{Jeremy BOUCHE-PILLON\affmark{a,$\ast$},  Pascale ZARATE\affmark{a,b}, Yannick CHEVALIER\affmark{a,c}  and Nathalie~AUSSENAC-GILLES\affmark{a}}

\affil{\affmark{a} Université de Toulouse - CNRS - IRIT, France}
\affil{\affmark{b}Université Toulouse Capitole, France}
\affil{\affmark{c}Université Toulouse 3 Paul Sabatier, France}
\email{Jeremy.Bouche-Pillon@irit.fr [First Author]; Pascale.Zarate@ut-capitole.fr [S. Author];\\ Yannick.Chevalier@irit.fr [T. Author]; Nathalie.Aussenac-Gilles@irit.fr [Fourth Author]}

\thanks{\affmark{$\ast$}Author to whom all correspondence should be addressed (e-mail: Jeremy.Bouche-Pillon@irit.fr).}

\historydate{Received DD MMMM YYYY; received in revised form DD MMMM YYYY; accepted DD MMMM YYYY}

\begin{abstract}

The emergence of the digital transition brought an increasing need to control the processing of digital information, including in Law Enforcement Agencies (LEAs). At the EU level, in recent years, many regulations have emerged to control data processing and exchange. Texts other than the GDPR, such as the ”Law Enforcement Directive (LED)”, appeared to regulate specifically how Law Enforcement Agencies  (LEAs) could process data. A formal representation of these regulations can be part of decision systems that support LEAs in processing data in compliance with the regulations. Although many new formalisms have emerged to represent legal norms and rules, few are provided with a reasoning mechanism. Furthermore, systems used in decision-making processes in critical contexts such as medical diagnoses or legal decisions cannot be fully automated, and the explainability of their results is essential to ensure user confidence in decisions.
This explainability aspect, while crucial, is lacking in most modern approaches that rely on machine learning. This paper describes a framework to operate formal rules from regulations, by focusing on explainability of the decision. \nouveau{After describing the general architecture of the proposed decision support framework, the paper showcases how symbolic AI and the SPARQL query language can support legal reasoning. It then describes an algorithm to generate a justification for the reasoning results, and outlines the procedure to be followed when the reasoning does not lead to a satisfactory conclusion. We notably focus on a method based on decision trees to determine what additional information to request from the user.}
\end{abstract}


\maketitle

\section{Introduction}\label{sec:Introduction}

Digital technologies present enormous growth potential. They promise to provide better and more seamless service and market access to citizens and economic agents by replacing existing services with digital ones that implement best-of-class solutions, are interoperable, and support  agents in their tasks. To be accepted, this digital transition\footnote{Digital transition - Reforms and Investments - European Union: \url{https://reforms-investments.ec.europa.eu/technical-support-instrument-0/digital-transition_en}} requires comprehensive governance to prevent the misuse of the introduced technology. Governance is often domain-specific and is defined through laws establishing guidelines and barriers on what can and cannot be done in finance, e-Health, administration, etc. For example, while the GDPR\footnote{Regulation - 2016/679 - EN - gdpr - EUR-Lex - European Union: \url{https://eur-lex.europa.eu/eli/reg/2016/679/oj/eng}} applies to all digital processing of EU citizens' personal data, the Digital Service Act (DSA)\footnote{Regulation - 2022/2065 - EN - DSA - EUR-Lex: \url{https://eur-lex.europa.eu/eli/reg/2022/2065/oj/eng}} applies more specifically to personal data processed in marketplaces and social networks. 

These regulations force agents willing to transform their activities to prove that their processes comply with laws whose interpretation may vary.
A solution is to introduce a Decision Support System (DSS) in which a qualified human agent is responsible for the description of the task at hand, and the system suggests a proper course of action to ensure compliance. To ensure that the human agent understands the legal implications, this advice should be supported with a legal reasoning justification based on the existing regulations. Such a DSS requires the formalization of regulations and reasoning on this formalization, as well as an appropriate communication with the human agent. 

Building such a DSS is challenging, as individual components themselves need to solve complex problems, and the provided solutions need to be compatible across the system. For example, many formalisms have been proposed to represent legal norms and rules, the prime examples being LegalRuleML~\citep{Palmirani:2011} and LKIF~\citep{Gordon:2008}.
However, this formalization needs to be operational, and to that end needs to be accompanied by a working mechanism for reasoning on the rules. For instance, LegalRuleML did not originally provide mechanisms to reason over the rules and supplementary works had to be carried out to allow it~\citep{Lam:2018}. The Carneades argumentation system~\citep{Gordon:2008}, designed to work with LKIF rules, could not be used successfully when tested and doesn't appear to be maintained, which shows the need for a standard-based solution. 

Another major issue, once the rules and the situation are described in a suitable logic, is the inability of pure logic reasoning mechanisms to explain the decision, only stating whether a decision is a logical consequence of its premises. That is, the pure reasoning should be complemented with a module that reflects upon the process that led to the result to provide the user with meaningful information. 

Finally, the growing complexity of the relations between stored pieces of information led to represent them using a standard format from Semantic Web technologies.
Instead of relying on \textit{ad hoc} relations stored in relational database tables, this framework promotes the use of well-defined entities and binary relations from standardized \emph{ontologies} to store information in \emph{Knowledge Graphs} (KG) as defined in~\cite{Lazarska:2019, Medhi:2017}, and provides the ability to query these graphs using standard query languages that are broadly supported with tools.

Based on all these observations, we propose to support decision making when one has to check the conformance of a situation to a set of rules describing one or more regulations.
\nouveau{In this paper, we will notably focus on establishing a method for generating explanations of the decision suggested. A state of the art on decision support systems presented in a previous paper~\citep{BP:These} led us to consider a symbolic approach, as opposed to learning-based approaches, given the critical nature of legal compliance checking. This approach also requires the development of methods for formally representing regulatory knowledge, legal rules and notions. The aim of this work is to answer the following research questions: (RQ1) How can symbolic AI be used to reason over formalized legal texts to verify compliance and provide justifications ? (RQ2) How to integrate semantic web standards in legal reasoning through the formalization of regulations? (RQ3) How to ensure that the results from legal reasoning provide a justified suggestion to the user? (RQ4) How to identify the most relevant information to ask a user when faced with an undecided scenario ?}

We propose to answer these questions by designing a framework that takes as input a situation description by a human agent using a specific ontology, and that supports reasoning on this situation thanks to formal rules represented using the Semantic Web's standard SPARQL. \nouveau{This part raises RQ1 and RQ2.}
If a decision can be reached, relevant legal sources are provided to support it. In the event of a failure, which occurs either in case of an inconsistency in the applicable rules or if the situation is insufficiently described, this reasoning module is complemented by an explainability module. This second module can analyze the cause of the failure and then ask the user for any missing information, or report any inconsistency met. \nouveau{Setting up this module requires to answer RQ3 and RQ4.}

To illustrate the design and use of this framework, we selected a use case about checking the conformance of data sharing and processing in LEAs to several European regulations. The central piece of legislation is the ”Law Enforcement Directive" (LED)\footnote{2016/680 - EN - Law Enforcement Directive; LED - EUR-Lex: \url{https://eur-lex.europa.eu/eli/dir/2016/680/oj/eng}}, which aims at regulating the processing of data related to investigations by Law Enforcement Agencies (LEAs) while respecting the right to privacy of EU citizens. This restricted scope makes this case tractable in terms of rule and situation complexity. In spite of this, it is rich enough to capture important concepts such as privacy, agents, processing, and the laws that can be stated over these.
Previous works have already covered the creation of the ontology~\citep{BP:2024_1} and of the formal rules~\citep{BP:2024_2} used in our studied use case. This paper focuses on developing an algorithm for generating an explanation to the rule-based reasoning results.

\paragraph{Outline} 
We first present the overall architecture of the framework and the relations between its components in Section~\ref{sec:framework}. Section~\ref{sec:rules} introduces the formalism used to represent rules taken from regulations. It is continued in Section~\ref{sec:reasoning} in which the different types of results that can be obtained through reasoning over these rules are presented, including the case of an undecided result.
This case is handled by the explainability module described in Section~\ref{sec:undecided}. In particular, since the undecided cases are caused by a lack of information in the description of the situation, this section presents the procedures employed to identify the supplementary information that should be asked of the user. This paper concludes in Section~\ref{sec:conclusion} with a summary on the presented works and a presentation of research perspectives.

\section{Decision-Support Framework}
\label{sec:framework}
Given the lack of explainable reasoning frameworks over formal legal rules, we aim to provide a DSS that interacts with the user to explain its decisions. To that end, we provide an interactive DSS by following the architecture described in~\citep{Marakas:2003} to obtain both the operability of formal rules and the explainability of reasoning results.

\subsection{General architecture}

\begin{figure}
  \centerline{\includegraphics[scale=0.55]{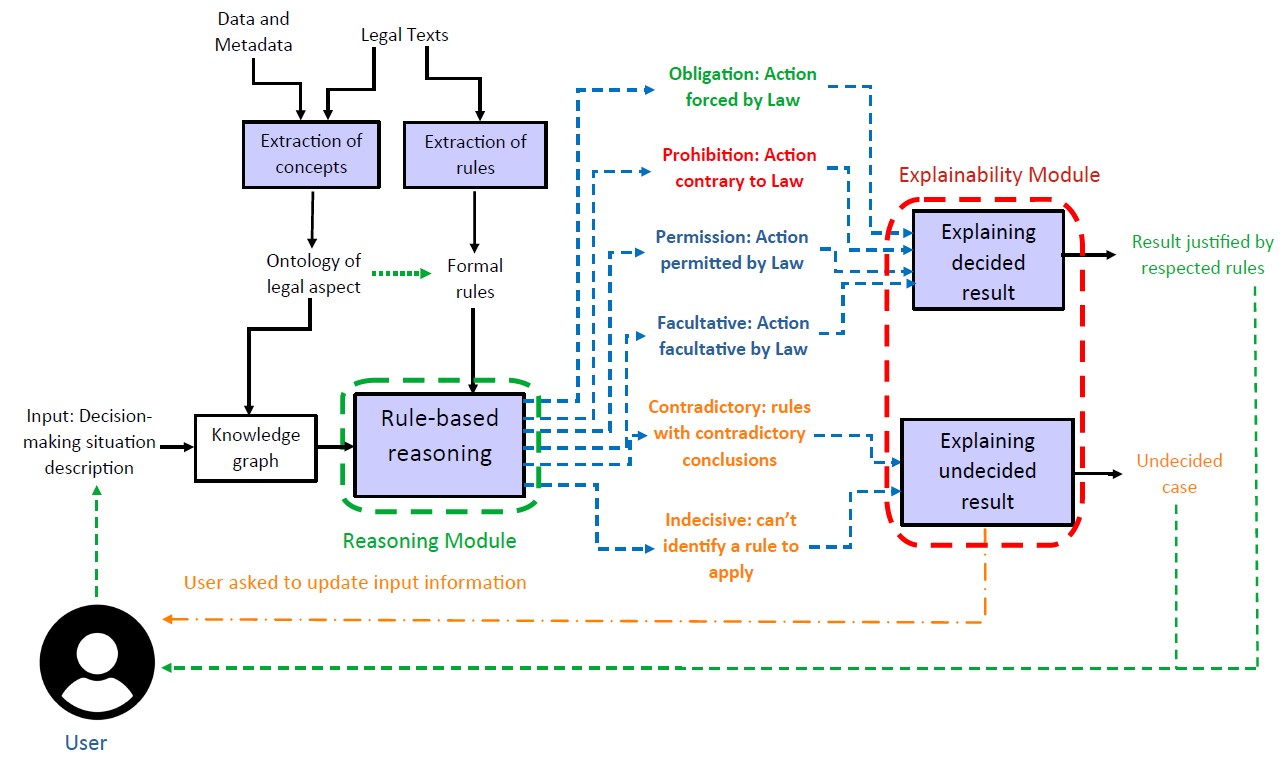}}
\caption{Framework principle\label{framework}}
\end{figure}

The DSS has been fully described in a previous paper~\citep{BP:2024_2}. Its main components 
are depicted in Fig. \ref{framework}. 

\begin{itemize}
    
    \item A first module, framed in green in Fig. \ref{framework}, implements a rule-based approach to ensure the respect of Law and regulations. One of the major drawbacks of rule-based systems is their inability to provide a response in cases where none of the rules in the system rule base is triggered. There is also a risk of triggering several contradictory rules. Based on this observation, it is already possible to envisage 2 different types of results for this first module:
    (i) Cases where the rules manage to reach a \textit{decided} conclusion, for example by stating that, \textit{"given the situation described in the input, an action is permitted"}. (ii) In the other cases, the rules do not manage to reach a satisfactory conclusion, and end up in an \textit{undecided} state.
    The handling of the result will differ depending on which of these two situations the first module ends up in.\\
   
    \item A second module, framed in red in Fig. \ref{framework}, is in charge of justifying the output of the reasoning module.
    While \textit{decided} results are easily justifiable, \textit{undecided} results do not provide a satisfactory answer. Therefore, it is advisable to improve the result before returning it is to the users, by asking them to adjust the input information so that the output could be \textit{decided}.
\end{itemize}

The operation of these modules is based on several other components that are also illustrated in Fig. \ref{framework} and described in the remainder of this section. 

\subsection{Legal sources}

\nouveau{The rule-based reasoning module relies on several components that contain information taken from legal sources such as laws and regulations. In most regulations, there are two types of information: (i) Definitions that clearly establish the terms and concepts involved, as well as the limitations of the regulations. These elements are integrated in the ontology and knowledge base, which will enhance the system's inference capabilities during the reasoning phase; (ii) The normative articles that use the terms and concepts defined earlier and establish legal rules. These rules are extracted and formalized to then serve as the basis of the reasoning module.}

\subsection{Ontology for the reasoning module}

Ontologies are a first-order logic vocabulary of entities and relations aiming at capturing the concepts in a given domain with a set of rules, and usually rely on the concepts described in existing ontologies. For example, a hierarchy of types of agents can be defined with the names of the types, a relation giving a type to agents, and rules capturing that agents in a subclass are also agents in the superclass. Their standardised description in RDF or OWL is the basis of the Semantic Web\footnote{Semantic Web: \url{https://en.wikipedia.org/wiki/Semantic_Web}}. Our DSS integrates a dedicated ontology to represent legal rules and situations in which a legal decision has to be computed. This ontology covers the representation of two types of legal knowledge:

\begin{itemize}
    \item The knowledge related to the legal rules and their metadata, like for example the deontic modalities they contain, or the legal documents they come from. This part of the ontology is generic and can be used for any legal compliance checking scenario. This part of the ontology can be built by extracting concepts and metadata from legal documents like regulations.

    \item The knowledge related to the specific type of situations the DSS is applied. It must capture the characteristics of the data, people, and authorities involved in the decision-making process, as well as contextual information, such as whether the decision is made in the context of an emergency. This part of the ontology is unique for each possible use case of the framework, and must be built specifically for the use case prior to using the framework. Building this part of the ontology  requires the analysis of the concepts related to the use case.
\end{itemize}

Accordingly, a situation is a \emph{knowledge graph} (KG), \textit{i.e.} a set of instances of the entities and relations in the ontology complemented with entities describing the situation. For example, a KG for our data-sharing use-case could have instances 'Alice \texttt{is a} \emph{LEA agent}' and 'Bob \texttt{is a} \emph{subject under investigation}' of the \texttt{'is a'} relation with the \emph{'LEA agent'} and \emph{'subject under investigation'} ontology-defined entities, and the situation-specific entities \emph{'Alice'} and \emph{'Bob'}. The rules of an ontology allows knowledge inference, that is, to reason over the most exhaustive description of a situation in a given use case. In the preceding example, instances stating that both Alice and Bob are agents are added to the set of instances. The encoded legal rules are also stored in a KG, and the rule-based reasoning module applies the rules stored in it to situations.
The ontology employed was presented in a previous article~\citep{BP:2024_1}, and the reasoning is described more precisely in the next three sections.

\subsection{Input of the reasoning module}

The input of the system consists of a description of the situation for which users want to check the conformance to legal rules.
All the data involved in the decision-making process are filled in by users using a form and then encoded in a knowledge graph (KG), which is where the reasoning takes place. The KG used in our framework fulfills several roles: It stores the ontology to allow semantic reasoning on the input data, but also the metadata about the rules used in the reasoning module. These metadata cover rule characteristics such as their deontic type or the legal document they come from; The KG can also be used to store data that are not part of the user's query, but that are part of the user's organization database and are also available to reason with; Finally it is used to store the input information when a user makes a decision support request to the system. While the ontology and the metadata about the rules are stored in a \textit{"global"} knowledge graph, two distinct approaches can be considered to store the input information:

\begin{enumerate}
    \item The input information can also be stored in the global KG, alongside the ontology and the rules metadata. This approach is simple to implement, but it raises the issue of user query independence. Indeed, if several queries are made on the system, there is a risk that some of them define instances of the ontology with the same identifier while being fundamentally different. This problem could lead to flawed reasoning and, therefore, result in incorrect decision suggestions. One way to avoid this issue would be to implement a system that guarantees the separation of entity identifiers between different queries made to the system.

    \item The other approach consists of storing each query in a specific named KG. This solution offers compartmentalization of the queries and prevents the reasoning process from being contaminated by external information unrelated to the query.
\end{enumerate}

Adopting one of these approaches influences the way the rules need to be written. Within the scope of the proposed framework, we selected the approach with named graphs, where compartmentalization of user queries prevents flawed reasoning by design.

\subsection{Formal rules of reasoning module}

The formal rules are extracted and derived from regulations and other legal sources, and their metadata are expressed using the ontology. The rules themselves are written in SPARQL, a language that allows to query the KG. The metadata are then inserted into the KG used by the DSS. To encode legal reasoning and be able to provide useful feedback to the user, we differentiate several types of rules:
\begin{itemize}
    \item Applicability rules that determine if legal articles apply to a situation, and allow for detecting when a situation is insufficiently described;
    
    \item Compliance rules that evaluate the compliance of a situation with legal articles;
    
    \item Exception rules that allow to define exception relationships between multiple rules, and thereby non-monotonic reasoning.
    
\end{itemize}

Each compliance rule is characterized by a deontic type (Permission, Facultative, Obligation or Prohibition) that is used in the reasoning process to evaluate the compliance of a situation with the regulations. It is important to note that the regulations from which the rules are extracted must, of course, be adapted to the scenario in which the framework is used.

\subsection{Possible outcomes of the rule-based reasoning module}

The first module, the \textbf{Rule-Based reasoning module}, develops a policy inspired by access control policies using formal rules extracted from regulations as well as concepts and relationships from the ontology. The two types of output, \textit{decided} and \textit{undecided} can be further divided using six output values. 

The \textit{decided} cases are divided in \textit{Obligation} (green text in Fig. \ref{framework}), \textit{Prohibition} (red text in Fig. \ref{framework}), \textit{Permission} and \textit{Facultative} (blue text in Fig. \ref{framework}). These types of results are obtained when several rules are complied with, and all arrive at the same deontic conclusion.

The \textit{undecided} cases (yellow text in Fig. \ref{framework}) can be either \emph{indecisive} or \emph{contradictory}. The reasoning is indecisive when no single formal rule from the rule base has been triggered. It is contradictory if several compliance rules are applicable but give contradictory deontic answers.
\nouveau{The handling of undecided cases calls for updating either the rule base or the input data, which occurs in the explainability module.}

\subsection{Explainability module}

The second main module of the framework generates a justification for the conclusion of the rule-based reasoning module. It is returned to the user alongside the decision suggestion. Explainability is indeed an essential aspect in decision support systems, and particularly if these systems are used in critical contexts, such as for example establishing medical diagnoses or judging legal cases. This module takes the results from the rule-based reasoning and depending on the type of result, generates an appropriate justification. 

\nouveau{In its current version, the proposed framework bases its explainability on the concept of \textit{'Enacted Law'}, as opposed to \textit{'Case Law'}. Rather than relying on the principle of precedence that would require an extensive database of previously determined cases, the justification of the decision comes from the statutes such as laws and regulations. Relying on \textit{'Enacted Law'} has two advantages in the scope of our study : First it provides a strict legal framework with which all institutions must comply, while \textit{'Case Law'} can differ between institutions. Second, the language used for \textit{'Enacted Law'} is very precise and defines the terms and their meanings for all parties affected by the law. By contrast, \textit{'Case Law'} generally cannot be captured by a single authoritative and uncontroversial formulation\footnote{IDENTIFYING APPLICABLE LAW - Law Explorer : \url{https://lawexplores.com/identifying-applicable-law/}}.}

Thus, in \emph{decided} cases, the conclusion of the reasoning module is presented together with the legal sources of the rules that led to the conclusion as a justification.

The \emph{undecided} cases however require special handling. \nouveau{In principle, \emph{contradictory} scenarios should not occur. While such cases may arise from contradictions in the applicable law, they would most likely be caused by wrongly encoded rules or by missing exceptions in the rule base. When faced with a \emph{contradictory} case, the system registers the conflict with the intention of forwarding it to an expert who will be able to determine how to resolve the conflict and update the rule base accordingly.}

For \emph{indecisive} cases, \nouveau{the system lacks information to reach a conclusion and will ask the user to update its input, while guiding him by indicating which information would lead to better chances of reaching a conclusion.} Section~\ref{sec:undecided} presents heuristics to identify information that, if added to the input, would allow the rule-based reasoning to reach a \textit{decided} conclusion.

We will now take a look at the different types of rules that intervene in the rule-base reasoning process and the formalism we chose to represent them.

\section{Using SPARQL for non monotonic reasoning}
\label{sec:rules}

\nouveau{The proposed framework is designed to be used in various situations and contexts where it is desirable to determine whether an action is permitted, required, prohibited, or optional. Depending on the context, it is necessary to consider the relevant laws and regulations in order to extract not only the underlying concepts but also the rules that will govern the decision-support process.}
The rules obtained that way are “explicit” and can be automatically extracted. Other \emph{implicit} rules, not explicitly present in regulations, are included to model “default cases” or exception rules defining when some rules should prevail over others.
Both the rules and the input contexts are represented as a graph using the concepts and properties of the ontology~\citep{BP:2024_1}. 

\nouveau{It was decided to use SPARQL to express the formal rules, as it allows to reason over RDF and OWL data. Some works conducted on the semantics of this language showcased its expressivity and showed that, subject to a few minor restrictions, it is equivalent to first-order logic. Notably, it is possible to translate the body of SPARQL queries into First-Order Logic (FOL) formulas using Answer Set Programming (ASP) as an intermediate language~\citep{Polleres:2013, Lee:2009}.}

The rules are applied using a SPARQL engine such as GraphDB\footnote{GraphDB homepage: \url{https://graphdb.ontotext.com}} or Apache Jena Fuseki\footnote{Apache Jena Fuseki homepage: \url{https://jena.apache.org/documentation/fuseki2/}} to reason directly over knowledge graph content. However, for clarity reasons, in this paper, the formalization will be presented in a FOL form following a $Conditions \to Effect$ syntax. The SPARQL equivalents of the FOL expressions can be found on our Github repository\footnote{Github repository: \url{ https://github.com/JeremyBOUCHEPILLON/legalDataProcessing/}} in the "rules" directory.

\subsection{Explicit rules from regulations}

Explicit rules are those that can be directly extracted from the regulations in natural language. Although extraction is currently manual, rule extraction from regulations can be automated using LLMs and semantic analysis~\citep{Fawei:2024, Ferraro:2020, Recski:2021}, which we plan to do in future works. The formalization principle applied in this framework is adapted from Gandon et al.’s work~\citep{Gandon:2017}, where each rule extracted from the regulation does not indicate directly whether an action is permitted, prohibited or mandatory. Instead each rule is classified as either permission, obligation or prohibition and the reasoning aims at assessing the compliance of the input situation to each rule. Moreover, as explained in a previous paper~\citep{BP:2024_2}, a legal rule is always encoded into two SPARQL queries, one indicating the applicability of the rule to the input situation, and the other indicating the respect of the rule by the situation. To illustrate this principle, the analysis of the article in Fig. \ref{article_example} would decompose as follows:

\begin{figure}
  \centerline{\includegraphics[scale=0.5]{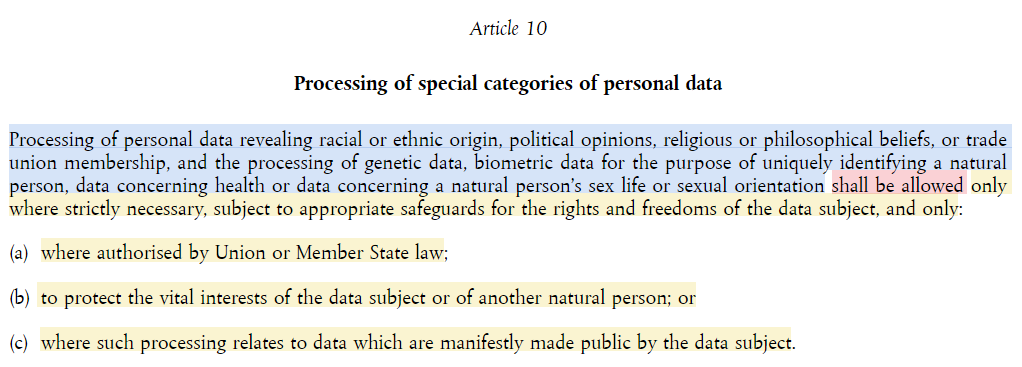}}
\caption{ Example of law article: article 10 from 2016/680/UE directive (LED)\label{article_example}}
\end{figure}

(i) The deontic class of this article is “Permission”, as indicated by the terms “shall be allowed”. (ii) The object of the rule relates to the processing of “sensitive personal data”. (iii) The conditions of this rule are “the strict necessity” of the processing, the “safeguard of rights and freedom of the data subject” and a disjunction of conditions: “allowed by Union or Member State” OR “protection of vital interests” OR “the data are public”. From this analysis, the “permission” aspect of the rule is expressed by the axiom \textit{isPermission(LED10)} and the rest can be formalized as the following 2 rules, in FOL:

\begin{itemize}
    \item $Processing(action) \land InvolvesDataset(action, dataset) \land ContainsData(dataset, data) \land SensitivePersonalData(data) \to IsApplicable(LED10, situation)$\\

    \item $IsApplicable(LED10, situation) \land Action(action) \land InvolvesDataset(action, dataset) \land Necessary(action) \land  SafeguardRights(action) \land (AuthorizedLaw(action) \lor ProtectsVitalInterest(action) \lor (\forall \, data, SensitivePersonalData(data) \implies PublicData(data))) \to HasCompliance(LED10, situation)$
\end{itemize}

\nouveau{Trying to write the second of these rules in SPARQL would give the query from listing \ref{SPARQL_LED10}}

\newpage 

\begin{small}
\begin{lstlisting}[language=SPARQL, caption=SPARQL query to express the compliance rule corresponding to the 10th article of the LED, label=SPARQL_LED10]
INSERT { graph ?g { nru:PS1 nrv:hasCompliance ?g }}
WHERE { 
	{graph ?g {
        ?action a :Processing .
        ?action :involvesData ?dataset .
    	?dataset :containsData ?data .
    	?data a :SensitivePersonalData .
    	?action :isNecessary "true"^^xsd:boolean .
    	{ ?action :isAuthorizedLaw "true"^^xsd:boolean }
    	UNION
    	{ ?action :protectsVitalInterests "true"^^xsd:boolean }
    	UNION
    	{ FILTER NOT EXISTS {
        	?data a :SensitivePersonalData .
        	FILTER NOT EXISTS {
            	?data a :PublicData .
        	}
    	}}
}}}
\end{lstlisting}
\end{small}

\nouveau{The query can be obtained through the automatic translation from First-Order Logic (FOL) formulas, applying the works conducted by \citet{Polleres:2013} and \citet{Perez:2009}, using Answer-Set Programming (ASP) as an intermediate language. It can be noted that this translation introduces \texttt{FILTER NOT EXISTS} to express both the classical logical negation and the existential quantifier’s negation. Since the universal quantifier does not exist in SPARQL, we go through the negation of the existential quantifier to express an equivalent formula: A formula of the form $\forall \, x , p(x)$ will be encoded in SPARQL as $\lnot \, \exists \, x , \lnot \, p(x)$.}

\paragraph{Substitutions} Consider a partial mapping \(\sigma\) from the set of variables \(\mathcal{X}\) to the set of terms. We extend it homomorphically to terms with \(\sigma(t)=t\) when \(t\) is not a variable.
Its \emph{domain} is the minimal set of variables \(D\subseteq \mathcal{X}\) such that \(\sigma(x)\) is defined. It is idempotent if for all \(x\in\mathcal{X}\) we have \(\sigma(\sigma(x))=\sigma(x)\). A \emph{substitution} is a homomorphic extension to terms (and then to relations) of an idempotent partial mapping on variables of finite domain. Substitutions are usually denoted in the postfix notation, \textit{i.e.} \(x\sigma\) rather than \(\sigma(x)\). We denote \(\Sigma_D\) a set of substitutions of domain \(D\), and given \(D'\subseteq D\), we denote \(\pi_{D'}(\Sigma_D)\) the set of substitutions that are a restriction to \(D'\) of the substitutions in \(\Sigma_D\). The \emph{join} of two sets \(\Sigma_D\) and \(\Sigma_{D'}\) of substitutions is denoted \(\Sigma_D \bowtie \Sigma_{D'}\) and is the largest set of substitutions of domain \(D\cup D'\) such that for all \(\sigma\in \Sigma_D\bowtie \Sigma_{D'}\) we have \(\pi_D(\sigma)\in\Sigma_D\) and \(\pi_{D'}(\sigma)\in\Sigma_{D'}\). The union of two sets of substitutions of common domain \(D\) is simply denoted \(\Sigma^1_D\cup\Sigma_D^2\). We note that there exists a unique set \(\Sigma_\emptyset\) that contains the substitution of empty domain and that it is a neutral element for the join.


\paragraph{Knowledge Graphs} The information encoded in a FOL model can be encoded as a \emph{Knowledge Graph} (KG) (within a larger domain and more predicates) with vertices representing entities and values, and arcs between 2 vertices labeled with a relation name. To bridge the gap with FOL models, a KG can also be seen as a unique relation between the head, the relation name, and the target of each arc. 
\begin{definition}{(Knowledge Graph)\label{def:KG}} 
    A \emph{Knowledge Graph} is a set of predicate instances \(\text{\tt PRED}(\text{\it const},\text{\it const},\text{\it const})\).
\end{definition}

\paragraph{SPARQLTree} SPARQL rules are of the form \(\text{\it body}\to \text{\it effect}\) where the \textit{body} designates a graph pattern formula. When applied on a knowledge graph \(K\), the effect is applied once for each substitution \(\sigma\) such that \(K\models \varphi\sigma\). To avoid the introduction of variables in \(K\), we assume that the set of variables occurring in the effect is a subset of \(X\).
Additional quantifiers can be used in defining \(\varphi^{X,W}\) through the \texttt{FILTER} construct. In the usual SPARQL syntax, a group \(\left\lbrace\varphi^{X,W} \text{\tt FILTER} \psi^{Y\cup X,W'}\right\rbrace\) has, as its set of solutions, the set of substitutions that are the restrictions on \(X\) of the substitutions of domain \(X\cup Y\) that satisfy both   \(\varphi^{X,W}\) and \(\psi^{Y\cup X,W'}\). This definition is not structural, which led us to the introduction of a structural version of SPARQL where the same construct is denoted \(\text{\tt FILTER}(\varphi^{X,W},\psi^{X\cup Y,W'})\). This change allows for easily defining what \(K\models \varphi\) means, and for the definition of mappings on SPARQL graph pattern formulas that will be developed \textit{infra}. We have omitted some constructions, such as equality or typing constraints, to focus on those used in this article. The grammar for this SPARQLTree query body is, denoting \(\text{\it t}\) the terms that are either variables or (literal) constants and decorations thereof:
\[
\begin{array}{rl}
\text{\it pred} & ::= \text{\tt PRED}(\text{\it t},\text{\it t},\text{\it t})\\ 
\text{\it body} & ::=
     \text{\tt AND}(\text{\it body},\text{\it body}) \,\vert\,
      \text{\tt UNION}(\text{\it body},\text{\it body}) \,\vert\,
       \text{\tt FILTER}(\text{\it body},\text{\it filter\_constraint})\\[1ex]
\text{\it filter\_constraint} &::= 
    \text{\tt NOT}(\text{\it filter\_constraint}) \,\vert\,
    \text{\tt EXISTS}(\text{\it body})\\
\end{array}
\]

\def\Variables{\ensuremath{\mathcal{X}}}
\begin{definition}
    The set of \emph{free variables} of a SPARQLTree expression \(\varphi\) is
    denoted \(\text{Var}(\varphi)\) and is defined inductively as
    \(\text{Var}(\text{\tt PRED}(h,p,t))=\lbrace h,p,t\rbrace\cap \Variables{}\), \(\text{Var}(\text{\tt AND}(\text{\it body}_1,\text{\it body}_2)) = \text{Var}(\text{\it body}_1)\cup\text{Var}(\text{\it body}_2)\), \(\text{Var}(\text{\tt UNION}(\text{\it body}_1,\text{\it body}_2)))= \text{\tt Var}(\text{\it body}_1) \cap \text{\tt Var}(\text{\it body}_2)\),and 
    \(\text{Var}(\text{\tt FILTER}(\text{\it body},\text{\it fc})) =
    \text{Var}(\text{\it body})\).
\end{definition}

The SPARQL language definition is intricate, and we have chosen here to present a simplified version to preserve clarity. For example, the definition of \texttt{UNION} means some variables may be unbound. For that reason, we have chosen to include in the free variables only those that must be bound in a solution. Beyond the implicit existential quantification on the variables of a query, nested universal and existential quantifiers are encoded using the \texttt{FILTER} construct, and thus variables introduced in the \texttt{EXISTS} part of the filter constraint are removed when computing the solutions.

Given a formula \(\text{PRED}(h,p,t)\) and a knowledge graph \(G\) we denote
\(s(\text{PRED}(h,p,t),G)\) the largest set of substitutions of domain \(\text{Var}(\text{PRED}(h,p,t))\) such that for every \(\sigma\in s(\text{PRED}(h,p,t),G)\) we have \(\text{PRED}(h,p,t)\sigma \in G\).

\begin{definition}{(Solutions of a SPARQLTree body)\label{def:KG:sol}}
    The application of a SPARQLTree formula \(\varphi\) on a KG \(G\)
    yields a set of substitutions \(\text{Sol}_G(\varphi,\Sigma_\emptyset)\)
    of domain \(\text{Var}(\varphi)\) defined inductively on \(\varphi\) as follows:
    \[
    \begin{array}{rcl}
       \text{Sol}_G(\text{PRED}(h,p,t),\Sigma_D) &=& \Sigma_D\bowtie s(\text{PRED}(h,p,t),G)  \\
       \text{Sol}_G(\text{\tt AND}(\varphi_1,\varphi_2),\Sigma_D) &=&  
        \text{Sol}_G(\varphi_2,\text{Sol}_G(\varphi_1,\Sigma_D))   \\
       \text{Sol}_G(\text{\tt UNION}(\varphi_1,\varphi_2),\Sigma_D) &=&   
       \pi_{(\text{Var}(\varphi_1)\cap \text{Var}(\varphi_2))\cup D}(\text{Sol}_G(\varphi_1,\Sigma_D) \cup
       \text{Sol}_G(\varphi_2,\Sigma_D))  \\
       \text{Sol}_G(\text{\tt FILTER}(\varphi,\text{\it fc}),\Sigma_D) &=& 
       \pi_{D'}(\text{Sol}^c_G(fc,\text{Sol}_G(\varphi,\Sigma_D)))\\
       \multicolumn 3r {\text{with }\ensuremath{D'}\text{ domain of }\ensuremath{\text{Sol}_G(\varphi,\Sigma_D)}}\\
    \end{array}
    \]
    The computation of solutions with constraints \ensuremath{\text{Sol}_G(\varphi,\Sigma_D)} is defined with:
    \[
    \begin{array}{rcl}
       \text{Sol}^c_G(\text{\tt NOT}(\varphi),\Sigma_D) &=& \Sigma_D\setminus \pi_D(\text{Sol}^c_G(\varphi,\Sigma_D))  \\
       \text{Sol}^c_G(\text{\tt EXISTS}(\varphi),\Sigma_D) &=& \pi_D(\text{Sol}_G(\varphi,\Sigma_D))  \\
    \end{array}
    \]
\end{definition}
A complete and more general computation of solutions is given in~\cite{Polleres:2013}, but Definition~\ref{def:KG:sol} is the basis of the algorithms provided \textit{infra} aiming at explaining the lack of solutions of a query on a KG \(G\).

\nouveau{In addition to the explicit rules extracted from regulations, we also need to integrate in our rule base potential default cases, expressed implicitly in legal texts. We also have to integrate conflict-solving rules to implement the defeasible aspect of legal reasoning.}

\subsection{Identifying implicit rules and conflict-solving rules}

In addition to the explicit rules, it is necessary to make explicit, with the help of legal experts, rules that are either implicit or that allow to solve conflicts between other rules. Those conflicts can have several causes. For example one rule may be an exception to another, or a legal doctrine defining priority between two rules may have not been applied (lex superior, lex posterior. This can be illustrated with the article in Fig. \ref{article_example}. Once analyzed, this article states that “the processing of sensitive personal data is permitted ONLY where some conditions are met”. Intuitively, this suggests that there is an implicit “default” rule stating that “the processing of sensitive personal data is forbidden” and the rule from the LED is an exception to this default rule. The formalization of  “default” rule is based on the axiom isProhibition(LED10\_default) and the FOL expression:

$Processing(action) \land InvolvesDataset(action, dataset) \land ContainsData(dataset, data) \land SensitivePersonalData(data) \to HasCompliance(LED10\_default, situation)$

And the rule that resolves the conflict states that if both rules are complied with, only the exception has to be considered (represented by negating the compliance with the default rule here):

$HasCompliance(LED10\_default, situation) \land  HasCompliance(LED10, situation) \to \neg HasCompliance(LED10\_default, situation)$


\nouveau{We therefore have a representation of legal rules based on the Semantic Web, thereby addressing our research question RQ2. The explicit rules form the core of our rule base and are direct formalizations of normative legal articles, while the implicit rules define default cases and the conflict-solving rules implement the defeasible aspect of legal reasoning. To justify the use of SPARQL, we have formally defined the elements of tree-based SPARQL syntax, SPARQLTree, noting that translations between FOL and SPARQL are possible via ASP. }

\section{Reasoning and explainability}
\label{sec:reasoning}

The extraction of explicit rules and the identification of implicit, exceptions and priority solving rules generate a full rule base on which to reason. The reasoning itself is decomposed into two steps: 
\begin{enumerate}
    \item All the explicit and implicit rules are applied on the situation KG, and if the premises of a rule are satisfied, its conclusion and an indication that it was satisfied are added to the situation KG; 
    \item All the exception and priority rules are applied on the obtained KG to solve potential conflicts by removing the decisions of the rules that are superseded by other rules.
\end{enumerate}

\nouveau{The reasoning aims not only at checking the compliance of a case with the regulation and at suggesting a decision to the end user, but it will also produce an explanation of this decision. This module of the framework deals with RQ3.}

\nouveau{At this point, the framework possesses a list of legal rules that apply to the input situation. The legal texts to be formalized within the framework are laws and regulations that specify what can and cannot be done in a given context. Numerous studies have examined the use of deontic logic to formally express this type of text~\citep{Jones:1992, Navarro:2014}. Although some works have explored complex relationships between different deontic modalities~\citep{Moretti:2009}, we base our study on a standard “deontic square” comprising four modalities that can be linked by a negation relation: The Permission (P) and its opposition the Prohibition/Interdiction (I), the Obligation (O) and its opposition the Facultative (F). Each normative rule formalized in the framework can be classified into one of these four categories.}


\nouveau{Thus, the list of satisfied rules generated by the reasoning module also provides the deontic modalities of those rules. In order to reach an overall conclusion regarding the situation entered by the user, it is necessary to understand how the various modalities interact, which will be detailed in the next section.}

\subsection{Evaluating the resulting deontic conclusion}

Establishing the resulting deontic conclusion from a set of respected rules requires to analyze all the possible combinations of deontic types.

To do that, we built a deontic truth table \ref{truth_table_deontic}, using the principles of the deontic square of oppositions~\citep{Moretti:2009}. It summarizes the relations between each combination of deontic types and the resulting global deontic conclusion, and its content reads as follows:
Each column represents a possible combination of deontic types, where \textit{1} denotes the presence of rules of this type in the list of respected rules, and \textit{0} the absence of rules of this type in the list. The letters \textit{F}, \textit{O}, \textit{I} and \textit{P} stand respectively for \textit{Facultative} (the right to not do), \textit{Obligation}, \textit{Interdiction} and \textit{Permission} (the right to do). In the result column, an $\emptyset$ represents a situation where no single rule is respected, and $\times$ highlights all the cases in which the respected rules lead to an inconsistent decision in deontic terms.\\

\begin{table}[h!]
    \caption{Truth table for the global deontic conclusion of a set of deontic rules \label{truth_table_deontic}}
    \centering
    \begin{tabular}{l|cccccccccccccccc}
        Case        &  0 & 1 & 2 & 3 & 4 & 5 & 6 & 7 & 8 & 9 & 10 & 11 & 12 & 13 & 14 & 15 \\\hline
        Facultative &  0 & 0 & 0 & 0 & 0 & 0 & 0 & 0 & 1 & 1 &  1 &  1 &  1 &  1 &  1 &  1 \\
        Obligation  &  0 & 0 & 0 & 0 & 1 & 1 & 1 & 1 & 0 & 0 &  0 &  0 &  1 &  1 &  1 &  1 \\
        Interdiction&  0 & 0 & 1 & 1 & 0 & 0 & 1 & 1 & 0 & 0 &  1 &  1 &  0 &  0 &  1 &  1 \\
        Permission  &  0 & 1 & 0 & 1 & 0 & 1 & 0 & 1 & 0 & 1 &  0 &  1 &  0 &  1 &  0 &  1 \\\hline
        Result      & \(\emptyset\)&P&I&\(\times\)&O&O&\(\times\)&\(\times\)&F&P and F&I&\(\times\)&\(\times\)&\(\times\)&\(\times\)&\(\times\)\\
    \end{tabular}
\end{table}

For example, Case 5 corresponds to a situation where some remaining decisions are Obligations, and others are  Permissions (there is a 1 in the \textit{O} and the \textit{P} columns, 0 in the others). In this situation, the deontic notion of obligation prevails over the permission, resulting in an Obligation result. In Case 3, the remaining decisions are of the Permission and Interdiction types (there is a 1 in the \textit{I} and the \textit{P} columns, 0 in the others). In this situation, interdiction and permission are conflicting deontic conclusions, and the computed result is a contradiction, denoted with the $\times$ symbol.

It can be noted that one specific combination in this truth table that combines only Permission and Facultative rules results in a double deontic conclusion, stating that the action is neither obligated nor prohibited (line 9). 
Another point of note is that Table~\ref{truth_table_deontic} distinguishes between Facultative and Permitted by making the former compatible with an interdiction and the latter compatible with an obligation.\\

The results from Table~\ref{truth_table_deontic} can be organized more succinctly in a Karnaugh-like map as in Table~\ref{karnaugh_deontic}. This representation allowed us to find a minimal simplified combination of deontic types to compute the conclusion of the rule-based reasoning module (\textit{cf.} logical formulas (\ref{eq_perm}), (\ref{eq_fac}), (\ref{eq_obl}), (\ref{eq_int}), (\ref{eq_conflict}) and (\ref{eq_empty})).

\begin{table}[h!]
\caption{Deontic Karnaugh table}\label{karnaugh_deontic}
\centering
\begin{tabular}{c c|c c c c} 
  & & \multicolumn{4}{c}{FO} \\
  & & 00 & 01 & 11 & 10 \\
 \hline
 \multirow{4}{1em}{IP} & 00 & $\emptyset$ & O & $\times$ & F \\
  & 01 & P & O & $\times$ & P \& F \\
  & 11 & $\times$ & $\times$ & $\times$ & $\times$ \\
  & 10 & I & $\times$ & $\times$ & I
\end{tabular}
\end{table}

\begin{align}
    P &= P \cdot \overline{I} \cdot \overline{O} \label{eq_perm}\\
    F &= F \cdot \overline{I} \cdot \overline{O} \label{eq_fac}\\
    O &= O \cdot \overline{I} \cdot \overline{F} \label{eq_obl}\\
    I &= I \cdot \overline{P} \cdot \overline{O} \label{eq_int}\\
    \times &= I \cdot P + F \cdot O + O \cdot I \label{eq_conflict}\\
    \emptyset & = \overline{P} \cdot \overline{I} \cdot \overline{O} \cdot \overline{F} \label{eq_empty}
\end{align}

The formula (\ref{eq_perm}) indicates that the resulting deontic conclusion is a permission if the list of respected rules contains permission rules and no interdiction and no obligation rules. Formulas (\ref{eq_fac}), (\ref{eq_obl}) and (\ref{eq_int}) are the analogous formulas for respectively facultatives, obligations and interdictions. The formula (\ref{eq_conflict}) indicates that there is conflict in three configurations: (i) There are Interdiction and Permission rules in the list; (ii) There are Facultative and Obligation rules in the list; (iii) There are rules classified as Obligation and Interdiction in the list.
Finally, the last formula (formula (\ref{eq_empty})) clearly expresses the fact that no rule is being respected, which forms the undecided case that is addressed \textit{infra}.

In conclusion, this deontic reasoning can thus generate 7 different output values that derive from 3 more general types of results. First, cases where at least one rule is satisfied by the input situation, with all the satisfied rules giving consistent decisions such that it is possible to draw a clear overall deontic conclusion (\textit{F}, \textit{O}, \textit{I}, \textit{P}, as well as \textit{P} \& \textit{F}). Secondly, cases where at least 2 rules are respected but their conclusions are inconsistent, creating a contradiction ($\times$). And lastly, undecided cases in which no compliance rule is satisfied by the input ($\emptyset$).
Each of these cases is handled in a different way, allowing to write an algorithm on how to handle the output of the reasoning mechanism.

\subsection{General algorithm for explainability}

The explanation given to the user will depend on the type of results from the reasoning phase, following the logic illustrated in Algorithm~\ref{alg:explain} \nouveau{that allows us to answer research question RQ3}. When all the respected rules are coherent, the explanation is straightforward and consists of informing the user of the regulation parts that support the result. However, two problematic cases may arise. The first  is when the rules produce contradictory decisions after the second conflict-solving step. This indicates that these rules are missing from the rule base. The result returned to the user highlights the contradiction, indicates the conflicting rules and asks him which rule should prevail. The answer is added as a new, temporary conflict-solving rule in the system, and is put forward for validation by an expert. The second problematic case arises when no compliance rule from the rule base has been respected. If no applicability rule is satisfied, the system indicates that the input situation is not in the scope of the verified regulations. The next section describes the cases in which some applicability rules are satisfied but no compliance rule. \\

\SetKwComment{Comment}{/* }{ */}

\begin{algorithm}[H]
\caption{Explanation of the rule-based reasoning results}\label{alg:explain}

\KwData{$(listApplicableRules, listRespectedRules)$ \Comment*[r]{Result from reasoning}}  
\KwResult{$(result, explanation)$ \Comment*[r]{The justified decision suggestion}} 

\uIf(\tcp*[f]{There are no applicable rules}){listApplicableRules is empty}{ 
    \Return “The situation is outside the scope of applicability of the verified regulations” \;
}
\uElseIf(\tcp*[f]{None of the applicable rules are respected}){listRespectedRules is empty}{
    $result \gets undecided$ \;
    \textcolor{red}{ask the user for potentially missing information to try to improve the result \;}
}
\Else{
    \uIf{all rules in listRespectedRules are Permission}{
        \Return (Permission, listRespectedRules) \;
    }
    \uElseIf{all rules in listRespectedRules are Facultative}{
        \Return (Facultative, listRespectedRules) \;
    }
    \uElseIf{all rules in listRespectedRules are either Permission or Facultative}{
        \Return (Permission \& Facultative, listRespectedRules) \;
    }
    \uElseIf{all rules in listRespectedRules are either Obligation or Permission}{
        \Return (Obligation, listRespectedRules) \;
    }
    \uElseIf{all rules in listRespectedRules are either Prohibition or Facultative}{
        \Return (Prohibition, listRespectedRules) \;
    }
    \Else(\tcp*[f]{There are contradictory respected rules}){
        \Return (Contradictory, listRespectedRules) \;
    }
}

\end{algorithm}

\section{Handling undecided cases using decision trees}
\label{sec:undecided}

In the case where no rule from the ruleset is respected, the system will ask the user for complementary information regarding the input situation. In order to determine which information to request from the user, the idea is to check what part of the rules that were “the closest to being complied with” were not respected. To do this, the following procedure is applied: (i) Consider only the rules “applicable” to the situation. (ii) For each of these rules, generate a binary decision tree where each node is one of the conditions of the rule, the left edge of a node corresponds to “the condition of the node is respected", and the right edge to “the condition of the node is not respected". (iii) Confront the input situation to each tree and keep a track of the non-respected conditions in them. (iv) sort the rules depending on how deep in their decision tree the verification went. (v) Use the non respected conditions in each rule to ask the user complementary information regarding the situation in input.

If the user is able to give supplementary information, the system processes the updated input from the beginning of the framework, hoping to obtain a \textit{"decided"} decision. After this new processing,if the reasoning fails again to find respected rules or if the user cannot give supplementary information, the case is definitely classified as \textit{“undecided”}.

\nouveau{The structure of the decision trees proposed in this section contributes to identifying relevant information to ask the end user, answering RQ4. In addition, the reasoning about undecided cases is an answer to RQ3 as it participates in providing better explanations to the end user.}

In the following section, we will detail the principle behind the tree construction (step 2 in the procedure), and then we will explore two approaches for confronting the input situation to the trees (step 3 of the procedure).

\subsection{Architecture of the decision trees}

The idea behind the constructed trees is to use their traversal to check the conditions of each applicable rule individually and establish which specific conditions were not fulfilled during the reasoning phase.

Each tree is a binary tree and implements an applicable rule. Each node  corresponds to one of the rule conditions, with the root node representing the first condition. Left edges correspond to a positive evaluation of a condition, while the right edges correspond to its negative evaluation. Each leaf on the tree may represent one of two types of results: (i) A positive output would indicate that all the rule conditions have been met for it to be respected. This type of leaf will never be reached during tree traversal because the construction of these trees is precisely motivated by the fact that no rule is being respected. (ii) A negative output contains the condition or the set of conditions that have not been met, explaining why the associated rule, although applicable, was assessed as not being respected during the reasoning process.

\paragraph{Analysis of failure} Def.~\ref{def:KG:sol} provides a total order of evaluation of the parts of a query. let us analyze the first failure according to the different cases with the notations of that definition (unrolling the cases of filter constraints):
\begin{description}
    \item[\texttt{PRED}]: Assuming \(\Sigma_D\neq\emptyset\) (first failure), the result is an empty set only if no instance of the predicate in \(G\) is joinable with a substitution in \(\Sigma_D\). In that case the explanation is the predicate;
    \item[\texttt{AND} and \texttt{UNION}]: A failure cannot occur at this level, it has to occur during the computation of either subformulas of \texttt{AND} or both subformulas of \texttt{UNION};
    \item[\texttt{FILTER EXISTS}]: This case behaves as an \texttt{AND} node with an added projection that cannot make the set of solutions empty;
    \item[\texttt{FILTER NOT EXISTS}]: This case results in an empty set of solutions if \(\Sigma_D\subseteq \pi_D(\text{Sol}_G(\varphi,\Sigma_D))\), \textit{i.e.} if \textbf{all} substitutions in \(\Sigma_D\) can be extended to satisfy \(\varphi\).
\end{description}

The last case often occurs in the translation of universal subformulas, and in that case \(\varphi\) is itself of the form \(\text{\tt FILTER}(\varphi_1,\text{\tt NOT}(\text{\tt EXISTS}(\varphi_2)))\). To provide a better explanation in that case, we introduce a new node \texttt{UNIVERSAL} of arity 2. Noting that a first-order logic formula \(\varphi_1\wedge(\forall x, \varphi_2 \Rightarrow \varphi_3)\)
is translated into \(\text{\tt FILTER}(\varphi_1,\text{\tt NOT}(\text{\tt EXISTS}(\text{\tt FILTER}(\varphi_2,\text{\tt NOT}(\text{\tt EXISTS}(\varphi_3))))))\), all such patterns are rewritten from the bottom-up into \(\text{\tt AND}(\varphi_1,\texttt{UNIVERSAL}(\varphi_2,\varphi_3))\).
By unrolling Def.~\ref{def:KG:sol}, we find that:
\[
\begin{array}{l}
\text{Sol}_G(\text{\tt UNIVERSAL}(\varphi_1,\varphi_2),\Sigma_D) \\
~~~~~~ = \text{Sol}_G(\text{\tt FILTER}(\top,\text{\tt NOT}(\text{\tt EXISTS}(\text{\tt FILTER}(\varphi_1,\text{\tt NOT}(\text{\tt EXISTS}(\varphi_2)))))),\Sigma_D)\\ 
~~~~~~ = \Sigma_D\setminus \pi_D(\text{Sol}_G(\text{\tt FILTER}(\varphi_1,\text{\tt NOT}(\text{\tt EXISTS}(\varphi_2))),\Sigma_D))\\ 
~~~~~~ = \Sigma_D\setminus \pi_D(\text{Sol}_G(\varphi_1,\Sigma_D)\setminus \pi_{D'}(\text{Sol}_G(\varphi_2,\text{Sol}_G(\varphi_1,\Sigma_D))))\\ 
~~~~~~ = \Sigma_D\setminus (\pi_D(\text{Sol}_G(\varphi_1,\Sigma_D))\setminus \pi_{D}(\text{Sol}_G(\varphi_2,\text{Sol}_G(\varphi_1,\Sigma_D))))\\ 
\end{array}
\]
Consider a substitution \(\sigma\in\Sigma_D\). It does not occur in the solution if it is present in  \(\pi_D(\text{Sol}_G(\varphi_1,\Sigma_D))\)---meaning extra variables can be instantiated to satisfy the condition of the filter---but not in \(\pi_{D}(\text{Sol}_G(\varphi_2,\text{Sol}_G(\varphi_1,\Sigma_D)))\)---meaning the conclusion of the universal cannot be satisfied with these additional instances. We note that failing to satisfy the condition \(\varphi_1\) of the universal is not a cause of failure. \textit{I.e.}, the explanation of the failure of a \texttt{UNIVERSAL} node is that of the failure of \(\varphi_2\). 

Let us now consider the origin of these rules. We assume that, as in our example, universal quantification is employed to constrain (using \(\varphi_2\)) the items described by \(\varphi_1\). These constraints are naturally a conjunction of properties that have to be satisfied. Instead of exploring \emph{why} a given property failed, we believe it is sufficient to handle the property as a whole. The same principle holds for the \texttt{FILTER NOT EXISTS} construct. We propose to handle this structure of explanation by constructing an \emph{explanation forest} to structure the search for the explanation of failure by adding sequence points in the sequential evaluation of the query.

\begin{definition}{(Explanation tree, explanation forest)\label{def:decision:forest}}
    An \emph{explanation tree} is either:
    \begin{itemize}
        \item A \texttt{NORMAL\_block} with a list of predicates;
        \item A \texttt{UNION\_block} with a list of explanation trees whose root is not a \texttt{UNION\_block};
        \item A \texttt{FILTER\_NOT\_EXISTS\_block} with two lists, one of explanation trees called its condition, and one of SPARQLTree formulas called its \emph{consequence};
        \item A \texttt{UNIVERSAL\_block} with a SPARQLTree formula called its \emph{condition},  and a list of SPARQLTree formulas called its \emph{consequence};
    \end{itemize}
    An \emph{explanation forest} is a list of \emph{explanation trees}.
\end{definition}

A SPARQLTree query body is translated into an explanation forest. Its construction from a SPARQLTree formula is given in the appendice\textbf{/is routine(if not in the appendice)}.

\begin{definition}{(Decision tree)\label{def:decision:tree}}
    A \emph{decision tree} is either a \texttt{OK} node, a \texttt{KO} node, or a \texttt{F}(ormula) node labeled with a SPARQLTree formula
    with two sons labeled one with \texttt{YES} and the other with a text providing an explanation.
\end{definition}

\begin{figure}[ht]
\centering
\includegraphics[scale=0.45]{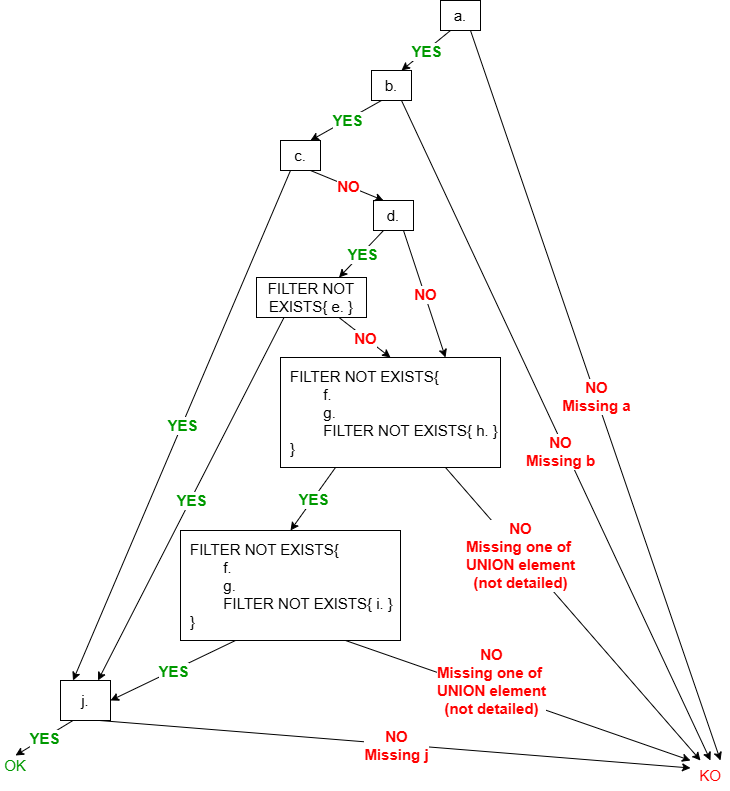}
\caption{Tree to determine the unfulfilled condition of the arbitrary rule from listing \ref{example_SPARQL_tree}}
\label{tree_exemple}
\end{figure}

Each explanation forest is translated into one decision tree with exactly one \texttt{OK} and one \texttt{KO} node. This transformation algorithm performs operations on nodes in trees, and we believe it is not particularly illuminating. We have presented a full translation example in Fig.~\ref{tree_exemple}, and described the translation patterns for each kind of explanation tree nodes in Fig.~\ref{fig:patterns}.\\

To illustrate what such a tree looks like, we will use simple arbitrary letters $a$, $b$, $c$, etc. to represent conditions or groups of conditions in the nodes. Starting from the root, recursively, the construction of the tree follows these principles:

\begin{itemize}
    \item  When the condition $a$ in the current node and the next condition $b$ are connected by a logical $\land$ operator (structure $a \land b$), the node containing $b$ is connected to the left branch leaving from the node $a$. In this case, it is because, if condition $a$ is not verified, then $a \land b$ as a whole is not verified. On the other hand, if $a$ is verified, in order to verify $a \land b$, it is still necessary to test whether $b$ is verified.
    \item When the condition $a$ in the current node and the next condition $b$ are connected by a logical $\lor$ operator (structure $a \lor b$), the node containing $b$ is connected to the right branch leaving from the node $a$.  In this case, it is because if condition $a$ is already verified, then $a \lor b$ as a whole is also verified. On the other hand, if $a$ is not verified, in order to verify $a \lor b$, it is still necessary to test whether $b$ is verified.
    \item The branch leaving from $a$ that is not yet connected is then connected either to a leaf or, if there are still other conditions connected with $\lor$ operators later in the rule, to the first node of the set of conditions that follows the block currently being processed.
\end{itemize}

Let's consider an arbitrary example of a simplified SPARQL query representing a rule with conditions expressed as simple predicates. Since the head is irrelevant for the construction of the tree, we will only consider the body of the query. 
In First-Order Logic, the rule would be written as $a \land b \land (c \lor (d \land \lnot e) \lor (\forall x. (f(x) \land g(x)) \implies (h(x) \land i(x)))) \land j$, which in SPARQL would be expressed as shown in listing \ref{example_SPARQL_tree}.  

\begin{small}
\begin{lstlisting}[language=SPARQL, caption=Body of a simplified SPARQL rule to illustrate the construction of the tree, label=example_SPARQL_tree]
a.
b.
{c.}
UNION
{
    d.
    FILTER NOT EXISTS{e.}
}
UNION 
{
    FILTER NOT EXISTS{
        f.
        g.
        FILTER NOT EXISTS{
            h.
            i.
        }
    }
}
j.
\end{lstlisting}
\end{small}

The query in listing \ref{example_SPARQL_tree} contains all the structures that can be found in a legal article: conjunctions and disjunctions of conditions, negations, universal quantification, some of them nested in each other.
In the corresponding SPARQL query, conjunction and disjunction are written  using respectively the dot and the UNION keyword. Universal quantification is encoded using the FILTER NOT EXIST construction.
For example, the second UNION of the query represents a condition \textit{"All the elements that verify f and g, are such that they also verify h and i"} that can be found in a legal rule "\textit{All the data that are personal and sensitive, are such that they also are public and disclosed by the person concerned}". In SPARQL, this universal quantification is expressed  with two negated existential operators, literally saying that \textit{"there is no f and g without also having h and i"} or, in the example, it can be rephrased as ("\textit{There is no data that is personal and sensitive, that is not also public and disclosed by the person concerned}"). A side-effect of rephrasing the universal quantification is to allow a more precise identification of the rule parts that could be violated.
The tree constructed from this rule and following these principles is shown in Fig. \ref{tree_exemple}.

\begin{figure}[ht]
\centering
\includegraphics[scale=0.45]{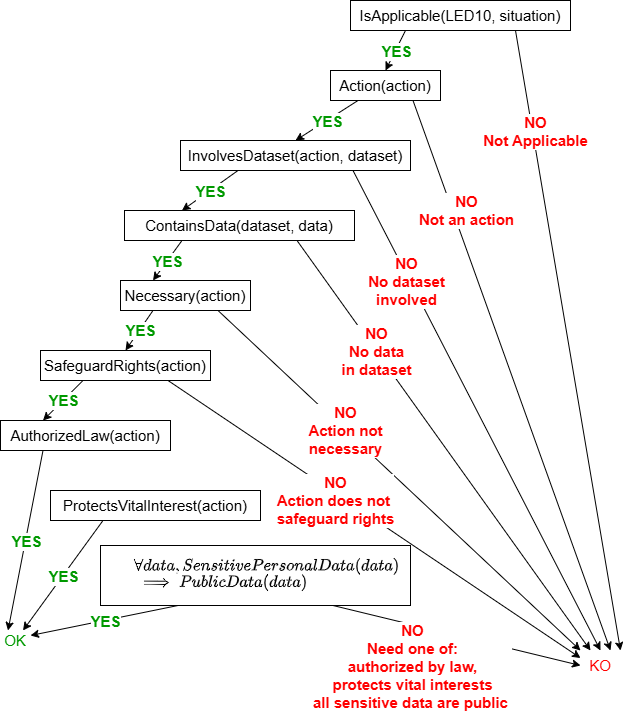}
\caption{Tree to determine the unfulfilled condition of the rule that checks for compliance with article 10 of the LED}
\label{tree_led10}
\end{figure}

If we apply the same principle to our running example of article 10 from the LED, we obtain the tree in figure \ref{tree_led10}.

\subsection{Constructing the trees from the SPARQL queries}

Now that we have established the principles behind the architecture of the decision trees, we need a procedure that allows us to transform SPARQL rules into this tree form.

First, the construction of the decision tree for each applicable rule requires parsing the related SPARQL rules about compliance.
This parsing is based on breaking down the SPARQL query into blocks of different types that can either be sequential or nested within each other.
The type of blocks may be \texttt{NORMAL\_Block}, \texttt{NEGATION\_Block}, \texttt{UNIVERSAL\_Block} and \texttt{UNION\_Block}, each of which is defined by different attributes:

\begin{itemize}
    \item The \texttt{NORMAL\_Block} is the most basic element to be parsed. It is defined by a list of strings, each of these strings being one of the triples of the query.

    \item The \texttt{NEGATION\_Block} is used when a part of the query is inside a simple \textit{"FILTER NOT EXISTS"} structure. It is defined by the list of the blocks inside the \textit{"FILTER NOT EXISTS"}.

    \item The \texttt{UNIVERSAL\_Block} allows to parse universal quantification in queries, that are expressed with two nested \textit{"FILTER NOT EXISTS"} in SPARQL. This block is defined by two lists of blocks. The first list corresponds to the part inside the first \textit{"FILTER NOT EXISTS"} of the structure and the second list to the elements inside the second \textit{"FILTER NOT EXISTS"} structure.

    \item The \texttt{UNION\_Block} represents all the alternatives of a \textit{"UNION"} structure in SPARQL. The datatype used to characterize it is a list of lists of Blocks. Indeed, a \textit{"UNION"} structure is composed of several alternatives, each of which is a sequence of blocks.
\end{itemize}

Finally, the query itself is parsed as a list of blocks.
To illustrate the parsing, let us consider the SPARQL structure from listing \ref{example_SPARQL_tree}. Parsing this example results in the following rule decomposition:

The query itself is composed of a succession of 3 main blocks, a \texttt{NORMAL\_Block} followed by a \texttt{UNION\_Block} and finally another \texttt{NORMAL\_Block}:

\begin{itemize}
    
    \item The first \texttt{NORMAL\_Block} contains the two conditions \textit{a} and \textit{b}.

    \item The \texttt{UNION\_Block} is composed of three alternatives:
    \begin{itemize}
        \item The first alternative is a simple \texttt{NORMAL\_Block} with the condition \textit{c}.

        \item The second alternative contains a \texttt{NORMAL\_Block} followed by a \texttt{NEGATION\_Block}:
        \begin{itemize}
            \item The \texttt{NORMAL\_Block} contains only the condition \textit{d}.

            \item The \texttt{NEGATION\_Block} contains a single \texttt{NORMAL\_Block} whose content is \textit{e}.
        \end{itemize}

        \item The third alternative contains a \texttt{UNIVERSAL\_Block}:
        \begin{itemize}
            \item The first part of the \texttt{UNIVERSAL\_Block} is composed of a single \texttt{NORMAL\_Block} that contains \textit{f} and \textit{g}.

            \item The second part of the \texttt{UNIVERSAL\_Block} contains also a single \texttt{NORMAL\_Block}, composed of the conditions \textit{h} and \textit{i}.
        \end{itemize}

    \end{itemize}

    \item The last \texttt{NORMAL\_Block} contains the last condition \textit{j}.

\end{itemize}

\begin{figure}[h!]

\centering
\begin{subfigure}{0.5\textwidth}
    \includegraphics[width=\textwidth]{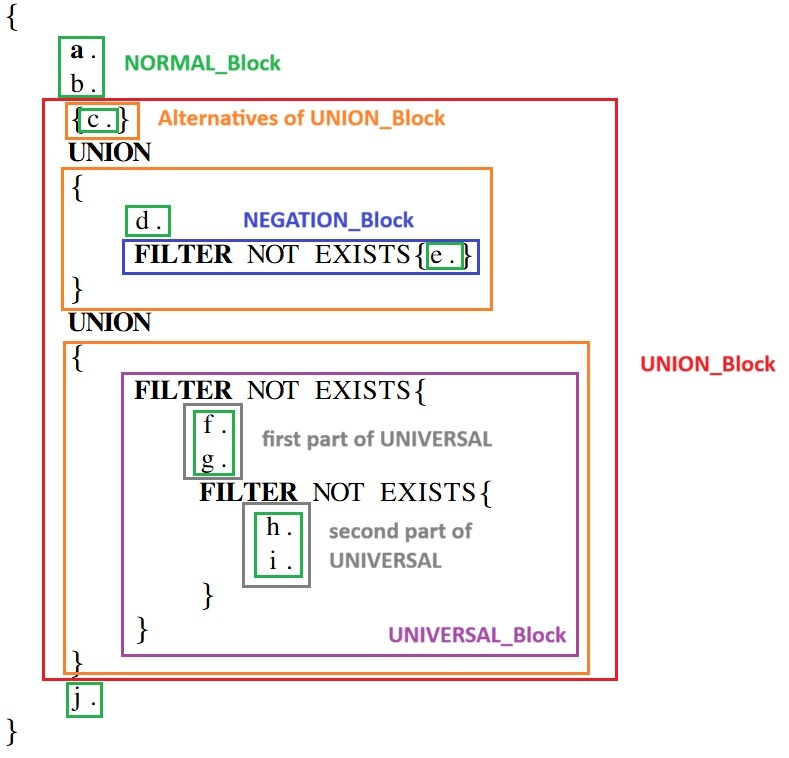}
    \caption{Blocks that compose the rule in the studied example}
    \label{parsing_visual}
\end{subfigure}
\begin{subfigure}{0.45\textwidth}
    \includegraphics[width=\textwidth]{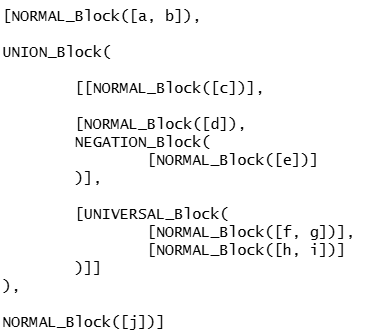}
    \caption{Textual translation of the parsing}
    \label{parsing_textual}
\end{subfigure}
\caption{Parsing of the blocks that compose the query used as example}
\end{figure}

Fig. \ref{parsing_visual} illustrates the different blocks that compose this example.
Block identification enables the construction of a decision tree  based on specific patterns that keep a one-to-one correspondence between the tree nodes and the rule structure and conditions.

\begin{figure}[ht]

\centering
\begin{subfigure}{0.30\textwidth}
    \includegraphics[width=\textwidth]{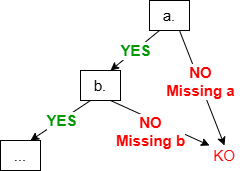}
    \caption{Pattern of tree building for a \texttt{NORMAL\_Block}}
    \label{normal_tree}
\end{subfigure}
\quad
\begin{subfigure}{0.30\textwidth}
    \includegraphics[width=\textwidth]{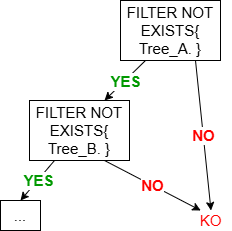}
    \caption{Pattern of tree building for a \texttt{NEGATION\_Block}}
    \label{negation_tree}
\end{subfigure}

\begin{subfigure}{0.45\textwidth}
    \includegraphics[width=\textwidth]{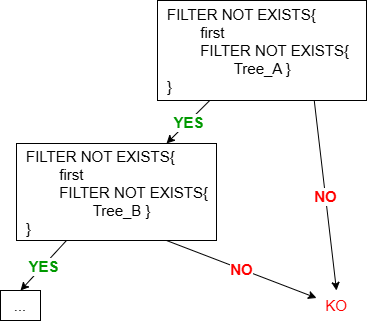}
    \caption{Pattern of tree building for a \texttt{UNIVERSAL\_Block}}
    \label{universal_tree}
\end{subfigure}
\quad
\begin{subfigure}{0.3\textwidth}
    \includegraphics[width=\textwidth]{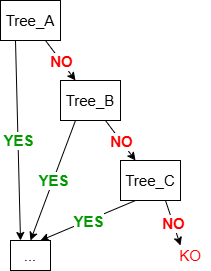}
    \caption{Pattern of tree building for a \texttt{UNION\_Block}}
    \label{union_tree}
\end{subfigure}

\caption{Patterns of tree building depending on the type of block parsed\label{fig:patterns}}
\end{figure}

\begin{itemize}
    \item The pattern to build a node from a \texttt{NORMAL\_Block} is the simplest. It consists of chaining the conditions in the block through left branches (the branches for a positive checking of the condition). An example for the block \texttt{NORMAL\_Block([a, b])} is given in Fig. \ref{normal_tree}.

    \item Building a node from a \texttt{NEGATION\_Block} only consists of prefixing the content of all the nodes constructed from the blocks inside the negation, as illustrated in Fig. \ref{negation_tree} for the example \texttt{NEGATION\_Block([Block\_A, Block\_B])}.

    \item Building a node from a \texttt{UNIVERSAL\_Block} is also straightforward, and consists of chaining the subtrees created from the blocks in the second part of the universal structure. The result of such construction from the example \texttt{UNIVERSAL\_Block(first, [Block\_A, Block\_B])} is given in Fig. \ref{universal_tree}.

    \item The pattern to build a node from a \texttt{UNION\_Block} is essentially the mirrored principle compared to a \texttt{NORMAL\_Block}. Each subtree created from the alternatives of the \texttt{UNION} are chained through the right branches (the branches for a negative checking of the condition). For example, the tree created from example \texttt{UNION\_Block([[Block\_A], [Block\_B], [Block\_C]])} is given in Fig. \ref{union_tree}.
\end{itemize}

Next, when going through the trees, it is necessary to establish how each node will be evaluated in order to take the correct branch at each step.
The evaluation of a node is actually carried out by executing a SPARQL query designed to check specifically the condition inside the node.
These queries are not INSERT queries like the ones used in the first module of the framework, but ASK queries. ASK is a SPARQL keyword that allows to test whether or not a query pattern has a solution, without returning information about the possible solutions. The binary response of these queries are then directly used as a way to determine which tree branch to take from each node. 
However, building the query pattern is challenging, as it requires to keep a track of the conditions that have already been checked as satisfied.

\begin{figure}[ht]
\centering
\includegraphics[scale=0.45]{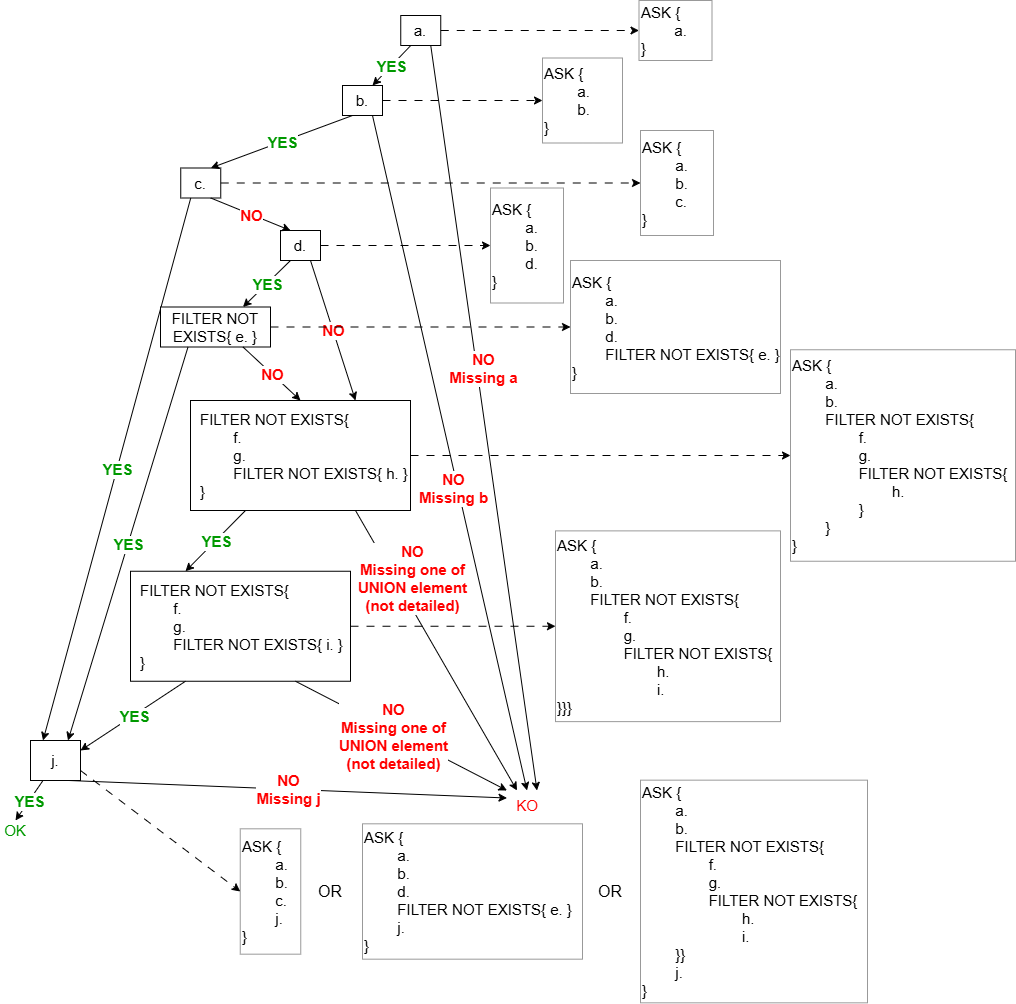}
\caption{SPARQL queries executed to check each node of the tree}
\label{example_tree_queries}
\end{figure}

Fig. \ref{example_tree_queries} gives an example of the queries constructed in each node of the tree in Fig. \ref{tree_exemple}.

We can note that three possible SPARQL queries can be executed in the last node. This is because prior to this node was a \texttt{UNION} with three alternatives. These alternatives are checked in order until one of them is verified. If one of the alternatives is verified, verification of others is skipped, and the conditions that are part of the verified alternative are registered in order to construct the SPARQL query for the following nodes.

When an \textit{undecided} conclusion is reached, a tree is constructed from each applicable rule in our system following the patterns presented. An algorithm then explores this forest to identify the conditions that are not actually met by the input situation.

\subsection{Exploring the forest to identify unfulfilled conditions}

Several approaches are possible for traversing the trees, in order to identify both the rules that were the closest to being respected and the unfulfilled conditions within these rules. We will explore two of them: (i) A first, naive approach, that looks only for the first set of decisive conditions that are not met in each applicable rule. (ii) A second, more optimized approach, which consists of searching for the largest subset of conditions that are satisfied by each applicable rule.

\subsubsection{A naive approach: Seeking the first unfulfilled decisive conditions}

On a concrete example, Fig. \ref{tree_led10} illustrates the tree that would be created from the compliance rule obtained from article 10 of the LED (cf. Fig. \ref{article_example}). For clarity and intelligibility, the nodes of this tree contain the conditions written in a more compact syntax, equivalent to what is written in SPARQL.
The first approach aims at identifying the first unfulfilled decisive conditions in each applicable rule to the input situation.
\nouveau{It can be noted that when the law articles are formalized, the conditions specified in the article, once translated, are listed in the same order in the SPARQL query than in natural language. This order tends to follow an order of increasing precision, where the further one goes into a article of law, the more requirements there are to verify, and the more detailed those requirements become.}
It is therefore relevant to consider that the further down the tree we go, the closer we are to fully complying with the rule.

We call \textit{decisive conditions} a subset of the conditions of the rules such that failure to meet all the conditions in this subset is necessary and sufficient for the rule to be violated. All leaves of the tree that are not a positive outcome (leaves that are not labeled \textit{"OK"} in our examples) contain one such subset. In other words, finding the first unfulfilled decisive conditions simply consists of finding the leaf reached when confronting a situation with the tree.

Given a tree created from a rule and a system input situation, the general principle is to go down the tree starting from the root node. For each node, take the branch that corresponds to the result of the binary check of compliance with the condition specified in that node. We stop when a leaf is reached, which gives us the first set of unfulfilled conditions of the rules.

Repeating the process for each tree created from the applicable rules gives us, for each of these rules, their set of unmet conditions as well as the depth of the leaf that was reached. From the depth reached and the maximum depth of the tree, we can calculate the relative depth reached, which can then be used as a metric to sort the rules in descending order of its value.

\nouveau{An algorithm that describes the implementation of this naive approach is provided in Appendix \ref{App:A}.}

However, this naive approach has two major flaws:

\begin{enumerate}
    \item It stops the validation of the conditions at the first unfulfilled decisive identified conditions, and doesn't check any further. By doing this, there is no guarantee that the information that the user is asked to provide will be sufficient to validate the rules. For instance, in our example, the procedure stopped after finding a missing information regarding the necessity of the action. But the end of the rule, with a disjunction of conditions, is not met either. With the naive approach, this lack will only be identified after the user has provided the required information about \textit{necessity} and the system has been run again.

    \item Using the relative depth reached is not an optimal metric either for ranking rules in order of their closeness to being satisfied. 
    For example, a rule can start with a part establishing a certain context, \textit{e.g.} that the user must be a police officer. That context is irrelevant in the evaluation of the rest of the rule, and a failure can just be the result of an unintended omission. However, with the naive approach, this rule would be put aside in terms of \textit{"closeness to being fulfilled"}.
    To formally express the issue, let us consider a situation that must be compared against two rules, both of which have the same number of conditions to check. The trees of these rules would then have the same maximum depth. For the first rule, let us assume that the situation meets all conditions except one, and this condition is very high up in the tree, among the first conditions checked.
    For the second rule, let us suppose that the situation meets almost all the conditions, except 3, but all these conditions are at the end of the tree, and are the very last conditions checked. According to our metric, the rule with the higher relative depth reached would be the second one, and yet, the first rule, with only 1 unfulfilled condition, should be considered \textit{the closest of being respected}.
    
\end{enumerate}

Although the naive approach proves to be sufficient in simple cases, the two flaws mentioned above make it ineffective in more complex cases, with rules involving numerous conditions to be checked to establish a proper context for the application of the rule.
This observation leads to the establishment of a more optimized strategy presented in the next section.

\subsubsection{An optimized approach: Seeking all the unfulfilled conditions}

This second approach aims at identifying not just the first, but all the unfulfilled decisive conditions in each applicable rule to the input situation. To do this, we need to check all the conditions of the rule, and note which ones are unfulfilled.

We achieve this by applying the following procedure to each rule:

\begin{enumerate}
    \item Construct the tree associated with the rule.

    \item Find and take note of the first unfulfilled conditions by applying the naive approach.

    \item Create a new version of the rule where the unfulfilled conditions have been removed. 
    Determining which part of the rule to remove is however not trivial. For example, if an unfulfilled condition is that a user is a member of a LEA, we should remove in the simplified version the dependent conditions that the LEA has jurisdiction over the place where the situation occurs, without removing the other conditions on that place.
    This step thus requires a dependency analysis among the conditions of the rule. The algorithm that carry out this analysis is described in details below.

    \item Create a new tree from the updated version of the rule.

    \item Check again for the first unfulfilled conditions.
\end{enumerate}

This procedure is repeated until the leaf reached in the tree is a leaf that signifies the rule is respected (a leaf labeled \textit{"OK"} in our examples).

By doing this, the initial rule gets split into two.
On the one hand, a rule that is respected, composed of a subset of the conditions of the initial rule. We call it the \textit{"Biggest respected sub-rule"}. On the other hand, there is the set of all the unfulfilled conditions of the rule, that can then be requested from the user all at once.

However, one aspect of this procedure requires special treatment.
Indeed, as mentioned in the third point of the procedure, when creating the new version of the rule where the identified unfulfilled conditions are removed, a problem of dependency between variables can occur. The conditions expressed in SPARQL are in the form of RDF triples that link a subject with an object through a specific property, with some of the elements being part of an ontology, and the others being variables. If, during the checking of the individual conditions of a rule, an unfulfilled condition is found and that this condition introduces a new variable in the SPARQL query, getting rid of the condition also suppresses the introduction of the new variable, and further conditions that depend on this variable need to be removed from the rule as well before generating the tree.
The difficulty lies in determining whether variables present in unfulfilled conditions are introduced into these conditions only.

To do so, we generalize our sequentiality assumption into a sequence-parallel assumption that the ordering on conditions is partial, but that along each path from the root new variables are introduced in relation with already determined variables.
We can then build a dependency graph, where nodes are the variables of the rule, and edges represent links that exist between these variables in the conditions of the rule.

Generally, let us separate the conditions of a rule in three groups. (i) The conditions already validated, (ii) the conditions that have just been evaluated as violated, and (iii) the conditions that have not been verified yet.
When creating the updated version of the rule, all the conditions in the second group will be removed. Then we need to determine whether conditions in the third group should be removed too. To do this, we consider each condition in the third group in turn. For each condition, we note the variables that appear in it. Using our dependency graph, we can determine whether all the variables in the current condition depend on variables that have been removed. If that is the case, the current unverified condition must be removed too. Otherwise, this means that there exists a link between the variables in this condition and the variables in validated conditions. In this case, we can keep it.

\section{Conclusions}
\label{sec:conclusion}

This paper presented reasoning and justification methods for a decision support framework. \nouveau{We notably focused on answering four research questions regarding the use of symbolic AI and semantic web for formalizing regulations, and the generation of justified suggestions, requiring, in cases where the reasoning fails to reach a conclusion, to identify the most relevant information to ask the user.} After describing the architecture of the framework and its components, we presented our approach formalizing legal rules. We then described the logic and algorithms behind the generation of a justified decision in our framework. Finally, to handle cases where not a single rule was satisfied, an innovative approach was presented to identify unfulfilled conditions in legal rules using decision tree structures. The use of the proposed framework has been illustrated in a use case of data sharing among European Law Enforcement Agencies (LEAs).
Several issues are still to be addressed in future works. 
First, the number of formal rules used to test our framework shall be extended, in order to see how it scales up to bigger use cases. This could be done by extracting rules from new directives that have come into force in 2023 regarding the data processing by LEAs\footnote{Directive - 2023/977 - EN - EUR-Lex - European Union: \url{https://eur-lex.europa.eu/legal-content/EN/HIS/?uri=oj:JOL_2023_134_R_0001}} and the gathering of electronic evidence in criminal proceedings\footnote{Directive - 2023/1544 - EN - EUR-Lex - European Union: \url{https://eur-lex.europa.eu/legal-content/EN/NIM/?uri=CELEX:32023L1544}}. We would also like to automate the extraction process of new formal rules, using state-of-the-art Natural Language Processing methods~\citep{Fawei:2024, Ferraro:2020, Recski:2021}. While the formal language used in this study is SPARQL, other formal standards like LegalRuleML~\citep{Palmirani:2011} could be tested to compare the results. \nouveau{Looking further ahead, we are also aware that the tests conducted on our framework are currently limited to proof-of-concepts, and we would like to establish a full benchmark to conduct large-scale experiments using real-world data. Finally, we plan to eventually expand our framework with a case-based reasoning module. This would complement rule-based reasoning by leveraging learning derived from past decision traces.}

\section*{Acknowledgments}

This work is partially funded by the H2020 project STARLIGHT (“Sustainable Autonomy and Resilience for LEAs using AI against High priority Threats”) that received funding from the European Union’s Horizon 2020 research and innovation program under grant agreement No 101021797. We would also like to thank Ronan PONS, PhD student in Law, who assisted this work by providing his insight as a legal expert.



\bibliographystyle{itor}
\bibliography{mybiblio_itor}

\vspace*{-10pt}
\appsection{Appendix A}\label{App:A}

\begin{algorithm}[H]
\caption{Tree traversal in the naive approach}\label{alg:naive}

\KwData{$query, tree, endpoint$ \Comment*[r]{Body of the SPARQL query, root of associated tree, endpoint to execute SPARQL queries}}

\KwResult{$(node, unrespected)$ \Comment*[r]{Node reached by traversing the tree, unrespected conditions}} 

$curr\_query \gets "ASK \{"$ \tcp*[f]{ASK query currently verified}

$node \gets tree$ \tcp*[f]{node initialized with the root}

$unrespected \gets []$

$temp \gets ""$ \tcp*[f]{temporary part of ASK query}

\While{node is not a leaf}{

\uIf(\tcp*[f]{content of UNION temporary}){node is the beginning of union sub-block}{
$temp \gets ""$

$temp\_query \gets curr\_query$

\While{node not the end of UNION sub-block}{
$temp \gets temp + node.content$ \tcp*[f]{adding condition in the node to temp}

$temp\_query \gets curr\_query + temp$ \tcp*[f]{adding temp to the query}

$result = ask\_query\_execution(temp\_query, endpoint)$ \tcp*[f]{boolean result}

$(node, unrespected) \gets next\_node(result, node, unrespected)$ \tcp*[f]{see Alg. \ref{alg:nextNode}}
}

\tcp*[f]{End of UNION sub-block}

$temp \gets temp + node.content$

$temp\_query \gets curr\_query + temp$

$result \gets ask\_query\_execution(temp\_query, endpoint)$

$(node, unrespected) \gets next\_node(result, node, unrespected)$

\uIf(\tcp*[f]{We finished traversing the whole UNION Block}){node is not the beginning of a new UNION sub-block}{
$curr\_query \gets temp\_query$ \tcp*[f]{the temp query becomes definitive}
}

}

\Else(\tcp*[f]{Not dealing with UNION}){
$curr\_query \gets curr\_query + node.content$

$result \gets ask\_query\_execution(curr\_query, endpoint)$

$(node, unrespected) \gets next\_node(result, node, unrespected)$
}      
}
\Return $(node, unrespected)$
\end{algorithm}

\begin{algorithm}[H]
\caption{next\_node: Function for advancing in the tree depending on the result of the ASK query}\label{alg:nextNode}

\KwData{$result, node, unrespected$ \Comment*[r]{boolean result of ASK query, current node, list of unrespected conditions}}

\KwResult{$(node, unrespected)$ \Comment*[r]{New node, updated list unrespected}}

\uIf(\tcp*[f]{ASK query returned true: condition verified}){result}{
$node \gets node.left$ \tcp*[f]{condition OK: left branch}
}
\Else{
add node content to unrespected

$node \gets node.right$ \tcp*[f]{condition NOK : right branch}
}

\Return $(node, unrespected)$

\end{algorithm}

\end{document}